\documentclass{article}

\usepackage{arxiv}

\usepackage[utf8]{inputenc} 
\usepackage[T1]{fontenc}    
\usepackage{hyperref}       
\usepackage{url}            
\usepackage{booktabs}       
\usepackage{amsfonts}       
\usepackage{nicefrac}       
\usepackage{microtype}      
\usepackage{lipsum}		
\usepackage{graphicx}
\usepackage[square,sort,comma,numbers]{natbib}
\usepackage{doi}
\usepackage{subcaption}
\usepackage{amsmath,amssymb,amsthm}

\usepackage{multirow}
\usepackage[bibliography=common]{apxproof}

\title{The SAME score: Improved cosine based bias score for word embeddings}

\author{Sarah Schröder \and \bf{Alexander Schulz} \and \bf{Barbara Hammer} \\
	CITEC, Machine Learning Group\\
	Bielefeld University - Faculty of Technology \\
	\texttt{\{saschroeder, aschulz, bhammer\}@techfak.uni-bielefeld.de} \\
}

\renewcommand{\shorttitle}{The SAME score}

\hypersetup{
pdftitle={The SAME score},
pdfsubject={cs.CL},
pdfauthor={Sarah ~Schröder, Alexander Schulz, Philip Kenneweg, Robert Feldhans, Fabian Hinder, Barbara Hammer},
pdfkeywords={Bias Detection, Fairness, Word Embeddings}
}

\newcommand{\extremaVgl}{magnitude-comparable} 
\newcommand{\minVgl}{unbiased-trustworthy}

\newcommand{\weatw}{$\textit{WEAT}_{\textit{sample}}$} 

\newcommand{\same}{\textit{SAME}}
\newcommand{\samew}{$\textit{SAME}_{\textit{sample}}$}

\newtheorem{definition}{Definition}[section]
\newtheorem{theorem}{Theorem}

\begin{document}
\maketitle

\begin{abstract}
With the enourmous popularity of large language models, many researchers have raised ethical concerns regarding social biases incorporated in such models. Several methods to measure social bias have been introduced, but apparently these methods do not necessarily agree regarding the presence or severity of bias. Furthermore, some works have shown theoretical issues or severe limitations with certain bias measures. \\
For that reason, we introduce SAME, a novel bias score for semantic bias in embeddings. We conduct a thorough theoretical analysis as well as experiments to show its benefits compared to similar bias scores from the literature. We further highlight a substantial relation of semantic bias measured by SAME with downstream bias, a connection that has recently been argued to be negligible. Instead, we show that SAME is capable of measuring semantic bias and identify potential causes for social bias in downstream tasks.

This is a preprint of the conference paper \cite{ijcnn24schroeder}. An earlier draft is available on arXiv \cite{samearxiv}.
\end{abstract}


\section{Introduction}
\label{sec:intro}

Over the last years, language models (LM) have become increasingly large and popular due to their superior performance and adaptability to many downstream tasks. Meanwhile many researchers have raised ethical concerns regarding social biases incorporated in such models. Plenty of methods for bias measurements have been introduced since, including some that measure semantic bias in word embeddings \cite{weat,bolukbasi}, methods that query the model intrinsic likelihoods \cite{crowspairs,nadeem2020stereoset,kurita2019measuring} or adaptions of common fairness measures \cite{biosbias,jigsaw} for downstream tasks. In recent years, there has been criticism about the focus on pretrained language models (PLM) and their intrinsic or semantic biases in general with the justification that these do not correlate with downstream bias or that testing conditions are too unreliable \cite{goldfarb2020intrinsic,seshadri2022templates,steed2022upstream}. To some extend this can be expected as there are many possible causes for biases \cite{shah2019framework}. At the same time, many methods to measure social bias are only motivated on an intuitive level and validated with limited empirical studies at best. However, some works have introduced benchmarks for bias score reliability \cite{du2021assessing,spliethoever2021bias} or theoretical criteria for bias scores \cite{icpram24schroeder}. In that context, it has been shown that WEAT \cite{weat}, a very prominent bias score for word embeddings, has limitations regarding bias quantification \cite{ripa,icpram24schroeder}. Prompted by these findings, we introduce \emph{Scoring Association Means of Word Embeddings} (SAME), a novel bias score for semantic bias in word or text embeddings. We use criteria set up in the literature to make certain that SAME is well suited to quantify semantic bias and validate its performance compared to similar bias scores in a thorough empirical analysis. We can show that it outperforms similar bias scores and does relate to downstream bias to a realistic extent considering that there are other causes for downstream bias. Our contributions are:

(i) We propose SAME, a new bias score for word or text embeddings;
(ii) We formally show that SAME has necessary properties for bias quantification introduced in \cite{icpram24schroeder};
(iii) We conduct experiments to support our theoretical findings and show that SAME outperforms existing bias scores in practice.

As opposed to WEAT, SAME can measure bias in terms of an arbitrary number of groups and it is more flexible as it allows to measure bias for a broad selection of words or texts without defining a hypothesis. It offers one aggregated score suited to compare biases between different models, while also offering interpretable bias components, which can be attributed to specific groups.
Our code is publicly available\footnote{https://github.com/HammerLabML/PLMBiasMeasureBenchmark/tree/ijcnn24}.

\section{Related Work}
\label{sec:related_work}

Cosine based bias scores are one method to measure semantic bias, i.e.\ "non-ideal associations between attributed lexeme (e.g.\ gendered pronouns) and non-attributed lexeme (e.g.\ occupation)" \cite{shah2019framework} of pretrained models, which is also referred to as intrinsic bias \cite{goldfarb2020intrinsic}. Semantic bias is one potential cause for biases in downstream tasks \cite{shah2019framework}. Thus different methods to measure semantic bias have been developed. We explain those that we compare SAME against. Furthermore, we highlight other works that evaluated bias scores and showed some limitations.
Using the notation from \cite{icpram24schroeder}, we will consider \textit{aggregated bias scores} (summarized over a set of target samples) and \textit{sample bias} (assigned to one word or sample), since both have application cases (group fairness vs.\ individual fairness).
Although our work focuses on semantic bias scores, we also go over some extrinsic ones (Section \ref{sec:exp_baseline_scores}), which we use in our experiments as comparisons.

\subsection{Cosine Bias Scores}

\subsubsection{WEAT}
The Word Embedding Association Test \cite{weat} 
was designed as a statistical test for stereotypes in word embeddings. Given two sets of target words $X$ and $Y$ and two sets of bias attributes $A, B$ of equal size $n$, the effect size which measures the association between those sets of words is defined as:
\begin{align}
\label{eq:weat_eff_size}
d(X,Y,A,B) = \frac{ \frac{1}{n} \sum_{\mathbf{x} \in X} s(\mathbf{x},A,B) - \frac{1}{n} \sum_{\mathbf{y} \in Y} s(\mathbf{y},A,B)}{stddev_{\mathbf{w} \in X \cup Y} s(\mathbf{w},A,B)},
\end{align}
with the association of a single word with the attributes
\begin{equation}
\label{eq:weat_attr_sim}
s(\mathbf{w},A,B) = \frac{1}{n} \sum_{\mathbf{a} \in A}\cos(\mathbf{w},\mathbf{a}) - \frac{1}{n} \sum_{\mathbf{b} \in B}\cos(\mathbf{w},\mathbf{b}).
\end{equation}
Under the hypothesis that words in $X$ are associated with attributes in $A$ and words in $Y$ with attributes in $B$, the effect size can confirm this. 
However, it cannot prove the absence of problematic associations, since its predictability is strictly limited to the hypothesis. Additionally, \cite{weat} 
propose a permutation test to measure statistical significance. However, we will focus on the effect size in terms of quantifying biases.

\subsubsection{gWEAT} 
A generalization of WEAT's test statistic for an arbitrary number of attributes is proposed in \cite{gweat}:
\begin{align}
    g(X_1,A_1, ..., X_n,A_n) = \sum_{i=1}^{n} (\overline{X_i} - \mu) (\overline{A_i} - \overline{A})
\end{align}
with $
    \mu = \sum_{i=1}^{n} \frac{\overline{X_i}}{n}$
assuming $n \geq 2$ (we do not consider $n=1$ in this work). It is not clear, whether they intend to apply this to embeddings normalized to length $1$, which would align with WEAT's usage of cosine similarity. However, we use it in that way, directly transferring WEAT's notation to $n \geq 2$ attributes.

\subsubsection{Direct Bias}
The Direct Bias by Bolukbasi et al. \cite{bolukbasi} measures the correlation of words $\mathbf{w} \in W$ with a so-called bias direction (in their notation the gender direction $\mathbf{g}$):
\begin{equation}
\label{eq:direct_bias}
DirectBias(W) := \frac{1}{|W|} \sum_{\mathbf{w} \in W} |\cos(\mathbf{w},\mathbf{g})|^c 
\end{equation}

where $c$ determines the strictness of bias measurement. The gender direction can be obtained by a single word-pair e.g.\ $\mathbf{g} = \mathbf{he} - \mathbf{she}$ or, 
it is obtained by computing the first principal component over a set of such difference vectors.

\subsection{Bias Score Evaluation}

Du et al.\ \cite{du2021assessing} investigate the reliability of bias scores for word embeddings under three criteria: Inter-rater consistency (do bias scores with similar definitions agree in their measurements?), internal consistency (are bias measures on individual samples of the same concept consistent) and test-retest reliability, (how much do bias scores vary when measured on different occasions, for example with different random data samples).
They found that the tested bias scores were not very consistent in their measurements. This aligns with the work of Seshadri et al.\ \cite{seshadri2022templates}, who found that bias scores are highly sensitive to templates in which target words were inserted to achieve more realistic bias measures for contextualized models.

In \cite{icpram24schroeder} two criteria to evaluate the usability of cosine scores in terms of bias quantification, e.g.\ for comparing embeddings after a debiasing intervention, are proposed. For $n$ protected groups represented by attribute sets $A = \{A_1$, ..., $A_n\}$ and a set of target words $T$, the criteria of comparability (Definition \ref{def:max_amplitutde}) and trustworthiness (Definition \ref{def:min_max_bias}) are defined. Comparability refers to a bias scores magnitude only depending on semantic bias as opposed to other structural properties of the model/ embedding space. Hence, this is a necessary criterion in order to compare biases in different models. Trustworthiness requires that a cosine score does not overlook biases, even if these only affect a minority of test samples. %

\begin{definition}[\extremaVgl \cite{icpram24schroeder}]
	\label{def:max_amplitutde}
	We call the bias score function $b(T,A)$ \extremaVgl\ if, for a fixed number of target samples in set $T$ (including the case $T = \{\mathbf{t}\}$), the maximum bias score $b_{max}$ and the minimum bias score $b_{min}$ are independent of the attribute sets in A:
	\begin{align}
	\label{eq:max_min_value}
	\max_{T, |T|=const} b(T,A) = b_{max} \; \forall \;  A, \quad  \\
	\min_{T, |T|=const} b(T,A) = b_{min} \; \forall \;  A.
	\end{align}
\end{definition}

\begin{definition}[\minVgl \cite{icpram24schroeder}]
	\label{def:min_max_bias}
	Let $b_0$ be the bias score of a bias score function, that is equivalent to no bias being measured.
	We call the bias score function $b(\mathbf{t},A)$ 
	\minVgl\ if 
	\begin{align}
	\label{eq:cond_neutral}
	b(\mathbf{t},A) = b_0 \iff s(\mathbf{t},A_i) = s(\mathbf{t},A_j) \; \forall \;  A_i, A_j \in A.
	\end{align}
	Analogously for aggregated scores with a set $T = \{ \mathbf t_1, ..., \mathbf t_m\}$,
	we say $b(T,A)$ is \minVgl\ if
	\begin{align}
	\label{eq:cond_neutral2}
	b(T,A) &= b_0 \nonumber \\
    \iff s(\mathbf{t}_k,A_i) &= s(\mathbf{t}_k,A_j) \; \forall \;  A_i, A_j \in A, k\in\{1,...,m\}.
	\end{align}
\end{definition}

They analyzed the Direct Bias and WEAT, showing that the Direct Bias and WEAT's effect size are not \minVgl\ and WEAT's sample bias $s(\mathbf{t},A,B)$ is not \extremaVgl.

\subsection{Extrinsic Bias Measures}
\label{sec:exp_baseline_scores}


\subsubsection{TPR Gap}
\label{sec:gap}
In \cite{biosbias} the true positive rate gender gap is introduced to measure gender bias in a downstream task, following the definition of Equalized Odds  \cite{hardt2016equality}, which is commonly used in the fairness literature. For one class $y$, the TPR Gap is defined as
\begin{align}
Gap_{g,y} &= TPR_{g,y} - TPR_{\sim g,y}, 
\end{align}
with $TPR_{g,y} = P[\hat{Y}=y | G=g, Y, y]$.

\subsubsection{Unmasking PLL}
\label{sec:rel_pll}
Nangia et al. \cite{crowspairs} propose to measure model intrinsic bias by the internal likelihood of a model to produce stereotypical content when certain groups are mentioned. More precisely they consider sentences, where a protected group is put into some stereotypical context, and a counterfactual sentence where that group is replaced by another group, which is less stereotypically associated with that context. In their example "John ran into his old football friend" vs. "Shaniqua ran into her old football friend", the modified tokens are \{John, his\} for the first sentence and \{Shaniqua, her\} for the second sentence.
Given one sentence version $S = U \cup M$, they assess the models likelihood $p(U|M,\theta)$ for the unmodified tokens $U$ (sentence context) given a group defined by the modified tokens $M$ (here male/female) and model parameters $\theta$. They found that they can best approximate this likelihood by the pseudo log likelihood
\begin{align}
    PLL(S) = \sum_{i=0}{|C|} log P(u_i \in U | U \setminus u_i, M, \theta).
\end{align}

A stereotype of the model would be confirmed for $PLL(S_{more}) > PLL(S_{less})$ if $S_{more}$ is the sentence with the stereotypical association and $S_less$ the counterfactual sentence. Hence, they compute a models bias as the percentage of samples with $PLL(S_{more}) > PLL(S_{less})$.

\section{SAME}
\label{sec:same}

To address the shortcomings of the existing cosine based bias scores, we propose a new score based on the before mentioned bias definition (Section \ref{sec:related_work}): \emph{Scoring Association Means of Word Embeddings} (SAME). It has a similar intuition to WEAT in terms of contrasting over attribute sets, but is meant to fulfill the criteria set up in \cite{icpram24schroeder}, as we will prove in Section \ref{sec:analysis_same}. Our empirical results in Section \ref{sec:experiments} support this. First, we define SAME in terms of a binary fairness problem, then extend to scenarios with multiple groups and finally show that it meets the criteria from \cite{icpram24schroeder}.

\subsection{Binary case}
\label{sec:same_binary}
We use one set of targets (typically words or sentences) $T$ and measure the association with two attribute sets $A_i$ and $A_j$. Furthermore, we assume that each attribute vector $\mathbf{a_i} \in A_i$ is normalized to unit length, so that $cos(\mathbf{t},\mathbf{a_i}) = \frac{\mathbf{t}}{|\mathbf{t}|} \cdot \mathbf{a_i}$. Then, the similarity of a word $\mathbf{t}$ towards an attribute set $A_i$ is
\begin{align}
\label{eq:sim_center}
s(\mathbf{t},A_i) = \frac{1}{|A_i|} \sum_{\mathbf{a_i} \in A_i} \frac{\mathbf{t}}{\|\mathbf{t}\|} \cdot \mathbf{a_i} = \frac{\mathbf{t}}{\|\mathbf{t}\|} \cdot \hat{\mathbf{a_i}} 
\end{align}
with $\hat{\mathbf{a_i}} = \frac{1}{|A_i|} \sum_{\mathbf{a_i} \in A_i} \mathbf{a_i}$. Similar to WEAT, we define a pairwise bias comparing the similarity of word $\mathbf{t}$ towards two attribute sets $A_i$ and $A_j$ with $\hat{\mathbf{a_i}} \neq \hat{\mathbf{a_j}}$:
\begin{eqnarray}
\mathbf{SAME}_{sample}(\mathbf{t}) &:=& b(\mathbf{t}, A_i, A_j) \notag \\
&=& \frac{s(\mathbf{t}, A_i) - s(\mathbf{t}, A_j)}{\|\hat{\mathbf{a_i}}-\hat{\mathbf{a_j}}\|} \notag \\
&=& cos(\mathbf{t},\hat{\mathbf{a_i}} - \hat{\mathbf{a_j}}).
\label{eq:same_pair_bias}
\end{eqnarray}
Contrary to \weatw\ (Equation \eqref{eq:weat_attr_sim}) we normalize the term, resulting in bias scores in $[-1,1]$ independently of the attributes and thus comparable. By transforming the equation, we can show that it has a similar notion to the Direct Bias from \cite{bolukbasi}, measuring the correlation of words with a bias direction between two attributes. However, we define the general bias direction as the mean of individual directions instead of using the PCA, avoiding the trustworthiness issue of the Direct Bias \cite{icpram24schroeder}, as will be shown in Theorem \ref{theor:own_min}.

To extend this score for an arbitrary number of words $\mathbf{t} \in T$, we take the mean absolute value of word biases:
\begin{eqnarray}
\label{eq:same_set_bias}
\mathbf{SAME}(T) &:=& \frac{1}{|T|} \sum_{\mathbf{t} \in T} |b(\mathbf{t}, A_i, A_j)|. \notag
\end{eqnarray}
The score is in $[0,1]$, independent of the attribute embeddings.

\subsection{Multi attribute case}
\label{sec:same_multi}
In order to apply SAME to a greater variety of fairness problems, we generalize it to cases with an arbitrary number of protected groups.
Let $A = \{A_0, ..., A_n\}$ with $n \geq 1$ contain at least 2 attribute sets.
To measure the bias with respect to all attributes in $A$, we construct a $n$-dimensional bias subspace from binary bias directions $\mathbf{\hat{a_i}}-\mathbf{\hat{a_0}}$ with $i \in \{1...n\}$ and $\mathbf{\hat{a_i}}$ the mean of attributes in $A_i$. Thereby, we assume that $\mathbf{\hat{a_i}} \neq \mathbf{\hat{a_j}} \; \forall i, j  \in \{1...n\}, i \neq j$.

Let $B$ be the bias subspace, defined by an orthonormal basis $\{\mathbf{b_1}, ..., \mathbf{b_n}\}$. The first basis vector $\mathbf{b_1}$ is obtained from the first binary bias direction, i.e.\
\begin{align}
    \mathbf{b_1} = \frac{\mathbf{\hat{a_1}} - \mathbf{\hat{a_0}}}{||\mathbf{\hat{a_1}} - \mathbf{\hat{a_0}}||}.
\end{align}
The other basis vectors are obtained from the successive binary bias directions, after removing linear correlations with previous basis vectors, which ensures orthogonality
\begin{align}
    \mathbf{b_i'} &= (\mathbf{\hat{a_i}} - \mathbf{\hat{a_0}}) - \langle \mathbf{\hat{a_i}} - \mathbf{\hat{a_0}} , \mathbf{b_{i-1}} \rangle \mathbf{b_{i-1}} - ... -  \langle \mathbf{\hat{a_i}} - \mathbf{\hat{a_0}} , \mathbf{b_0} \rangle \mathbf{b_0} \nonumber 
\end{align}
and $\mathbf{b_i} = \frac{\mathbf{b_i'}}{||\mathbf{b_i'}||}$. Thereby, we assume that $||\mathbf{b_i'}|| \geq 0$. In the case of linear dependency of the binary bias directions, i.e.\ one calculated $\mathbf{b_i'}$ is a zero vector, this vector can be left out, resulting in a smaller bias space $B$.

Given $B$, the bias of a target embedding $\mathbf{t}$ is described by the cosine similarities with the basis vectors of $B$
\begin{align}
    \mathbf{w_B} = \Big( \cos(\mathbf{t} , \mathbf{b_1}) \; , \; ... \; , \; \cos(\mathbf{t} , \mathbf{b_n}) \Big) ^T
\end{align}
and the bias magnitude is
\begin{align}
    SAME_{sample}(\mathbf{t}) := b(\mathbf{t}, A) = ||\mathbf{w_B}||
\end{align}

Consequently the aggregated bias of all target words $W$ is
\begin{eqnarray}
\label{eq:same_set_bias_multi}
SAME(T) := b(T,A) &=& \frac{1}{|T|} \sum_{\mathbf{t} \in T} b(\mathbf{t},A). %
\end{eqnarray}

A major benefit of using a vector $\mathbf{w_B}$ as an intermediate step to obtain a bias magnitude over multiple attributes is that we achieve an interpretable representation of $\mathbf{t}$ in terms of the protected groups, i.e.\ each element in $\mathbf{w_B}$ represents the association of $\mathbf{t}$ with the two groups, whose attributes were used to obtain the respective basis vector. Noteworthy, all elements represent bias in comparison to the same default group that is represented by $A_0$ and users must be aware that if bias directions correlate with on each other, their share will be only accounted for by the first basis vector. Therefor, analyzing how the different bias directions correlate with each other is highly recommended to allow more sophisticated insights into how biases manifest in embedding spaces.

\subsection{Analysis of SAME}
\label{sec:analysis_same}

The following theorems and their proofs detail properties of SAME in light of definitions \ref{def:max_amplitutde} and \ref{def:min_max_bias}. We show that \same\ and \samew\ are \minVgl\ and \extremaVgl\ for the multi-attribute case, 
which also covers the binary case. Hence, we can state that SAME is a reliable bias score to quantify bias in embeddings and it can be compared between different embedding models. 

\begin{theorem}
\label{theor:own_min}
The bias score function $b(\mathbf{t},A)$ and $b(T,A)$ are \minVgl .
\end{theorem}
\begin{inlineproof}
The proof is in the Appendix for better readability. 
\end{inlineproof}
\begin{toappendix}
\begin{proof}
\label{proof:own_min}
for \textbf{Theorem} \ref{theor:own_min}.
The bias score indicating no bias is $b_0 = 0$. First, we can state that
\begin{align}
b(\mathbf{t},A_i, A_j) = 0 \iff s(\mathbf{t},A_i) = s(\mathbf{t},A_j),
\end{align}
which directly follows from the definition. 
Hence 
\begin{align}
    b(\mathbf{t},A) &= ||\mathbf{w_B}|| = ||\big( b(\mathbf{t},A_0,A_1), ..., b(\mathbf{t},A_0,A_n) \big)^T|| = 0 \nonumber \\
    &\iff s(\mathbf{t},A_i) = s(\mathbf{t},A_0) \forall A_i \in A \nonumber \\
    &\iff s(\mathbf{t},A_i) = s(\mathbf{t},A_j) \forall A_i, A_j \in A \nonumber 
\end{align}
and 
\begin{align}
b(T,A) = 0 \iff b(\mathbf{t},A) = 0 \; \forall \mathbf{t} \in T.
\end{align}
\end{proof}
\end{toappendix}


\begin{theorem}
\label{theor:own_extrema}
	The bias score functions $b(\mathbf{t},A)$ and $b(T,A)$ are \extremaVgl, assuming the number of protected groups $n$ is not larger than the dimensionality of the embedding space.
\end{theorem}

\begin{inlineproof}
The proof is in the Appendix for better readability. 
\end{inlineproof}
\begin{toappendix}
\begin{proof}
\label{proof:own_extrema}
for \textbf{Theorem} \ref{theor:own_extrema}.
For SAME we now show that $b_{min} = 0$, $b_{max} = 1$ and both can be reached independent of $A$. Since $A$ defines the bias space $B$ used by SAME, we need to show that the extreme values can be reached independent of $B$. First, we can state that
\begin{align}
\max_T b(T,A) &= \max_\mathbf{t} b(\mathbf{t},A), \\
\min_T b(T,A) &= \min_\mathbf{t} b(\mathbf{t},A)
\end{align}
which is derived directly from the definition of $b(T,A)$.

With an embedding space in $\{\mathbf{d}_1, ..., \mathbf{d}_n\}$ the orthonormal basis for the bias space $B$, we can write any vector $\mathbf{t} \in \mathbb{R}^d$ as a linear combination of its parts $\mathbf{w_{\parallel B}} \in B$ and $\mathbf{w_{\perp B}} \not\in B$ and the former one as the sum of projections onto the basis vectors
\begin{align}
    \mathbf{t} = \mathbf{w_{\perp B}} + \mathbf{w_{\parallel B}} = \mathbf{w_{\perp B}} + \sum_{i} \langle \mathbf{t}, \mathbf{d}_i \rangle \mathbf{d}_i.
\end{align}
With $||\mathbf{b_i}|| = 1$ and $\mathbf{b_i} \perp \mathbf{b_j} \; \forall i,j  \in \{1,...,n\}, i \neq j$ and
\begin{align}
    \mathbf{w_B} =& \Big( cos(\mathbf{t} , \mathbf{b_1}), ..., cos(\mathbf{t} , \mathbf{b_n}) \Big)^T \nonumber \\
    =& \frac{1}{||\mathbf{t}||} \Big( \langle \mathbf{t}, \mathbf{d}_1 \rangle, ..., \langle \mathbf{t}, \mathbf{d}_n \rangle \Big)^T 
\end{align}
follows
\begin{align}
    &\implies& b(\mathbf{t},A) = ||\mathbf{w_B}|| = \frac{1}{||\mathbf{t}||} ||\mathbf{w_{\parallel B}}||.
\end{align}

With $||\mathbf{w_{\parallel B}}|| \leq || \mathbf{t}||$ we can determine the upper bound of the bias magnitude and the lower bound follows directly from the definition:
\begin{align}
    0 \leq b(\mathbf{t},A) \leq 1
\end{align}

To show that both extreme cases can be reached independent of $A$, we consider the following extreme cases:
First, let $\mathbf{t}$ be orthogonal to the bias space $B$, i.e. $\mathbf{t} = \mathbf{w_{\perp B}}$, which is possible as long as the bias space $B$ is lower dimensional than the embedding space. Then follows $||\mathbf{w_{\parallel B}}|| = 0 \implies b(\mathbf{t},A) = 0$. Since $B$ has $n-1$ dimensions for $n$ attribute groups, this requires that an embedding space with $d \geq n$ dimensions as provided in the statement.
Secondly, let $\mathbf{t}$ be entirely defined in the bias space, i.e. $\mathbf{t} = \mathbf{w_{\parallel B}}$. Then follows $||\mathbf{w_{\parallel B}}|| = ||\mathbf{t}|| \implies b(\mathbf{t},A) = 1$.
Hence the statement follows.

\end{proof}
\end{toappendix}

\section{Experiments}
\label{sec:experiments}
In our experiments we investigate how reliable cosine scores are in practice and how our proposed score SAME compares to other state-of-the-art scores. We outline our experiments scope and goals in \ref{sec:exp_goals}, introduce the datasets (Section \ref{sec:exp_datasets}) and describe the experiment setup (Section \ref{sec:exp_setup}).

\subsection{Experiment Scope and Goals}
\label{sec:exp_goals}
Cosine based bias scores were proposed to investigate biases in PLMs and thus should be able to reflect the biases manifesting in a pretraining task. However, in practice PLMs are part of a learning pipeline, i.e.\ being finetuned to a downstream task or being used to obtain embeddings, on which some classifier, clustering or ranking approach is build. To address both cases we evaluate the cosine scores in the scope of masked language modelling (MLM) and two downstream classification tasks. While it is common practice to finetune all layers of a language model, this influences intrinsic/semantic bias. Hence, we treat PLM (embeddings and MLM head) as static in our experiments.
To allow a fair comparison of semantic and extrinsic bias, we measure all bias scores on the same test data. 
In order to show the effectiveness of SAME compared to other cosine scores, we conduct experiments covering the following aspects:

\subsubsection{Discriminating more or less biased models}\label{sec:discr_models}
A benefit of cosine bias scores is that we can test PLM for biases without deploying them to some downstream task, which may involve costly training. However, this assumes that cosine scores can be used to identify models that would show more or less bias in a downstream task. In that case, they may also attest if debiasing on embeddings of the PLM will reduce biases downstream.
For that reason, we conduct pairwise comparisons of models that exhibit particular low or high downstream biases and test if aggregated cosine scores agree with the downstream bias. Specifically, we chose those models, whose downstream bias deviated by more than the standard deviation from the mean bias of all models.

\subsubsection{Correlation of cosine scores with downstream bias}\label{sec:corr_cos}
A stronger argument for the meaningfulness of cosine scores would be correlations with downstream biases using an arbitrary selection of models. However, as emphasized in the bias framework of \cite{shah2019framework}, downstream biases are influenced by other factors than the semantic bias alone. 
Thus the impact of semantic bias may be concealed by other factors such as biases in the task specific data. Hence, we expect limited correlations for all cosine scores. However, significant differences can still be interesting.
We test three cases: correlations for counterfactual biases (of individual samples), sample biases aggregated per class and aggregated biases over all test samples.

\subsubsection{Robustness of bias scores} 
Following \cite{du2021assessing} we measure the test-retest reliability of the different cosine scores in terms of selected targets samples. 
Specifically, we compute gender biases of jobs using the BIOS datasets. As baselines, we compute the aggregated bias scores on all job titles and all samples of the BIOS dataset (see \ref{sec:exp_datasets}). Then we compute the bias scores on random permuted subsets of the dataset and measure the absolute deviation of bias scores, normalized to the interval $[0,1]$ for better comparison. For WEAT we sort targets into sets $X$ and $Y$ based on the jobs distribution in the current test data. The amount of test data ranges from 20\% (the amount of test data used in our previous experiments) to 1\% from a total of $10602$ bios. The latter case is more similar to test cases in the literature, where words are inserted into a limited amount of templates to generate "more diverse" data.

\subsection{Datasets}
\label{sec:exp_datasets}

We use three downstream datasets developed to investigate biases in LMs: In terms of classification we consider the Jigsaw Unintended Bias (Jigsaw) dataset \cite{jigsaw} and the BIOS dataset \cite{biosbias}. For MLM we use the CrowS-Pairs dataset \cite{crowspairs}.

\subsubsection{JigsawBias}
The Jigsaw Unintended Bias datasets was introduced \cite{jigsaw} as an extension of the "Toxic Comment Classification Challenge" on Kaggle\footnote{https://www.kaggle.com/c/jigsaw-toxic-comment-classification-challenge/data}. It provides toxicity labels alongside labels for specific types of problematic content (e.g.\ threats) and labels for protected groups.
In our experiments, we use the dataset for binary classification (toxic / not toxic) and, to limit computational costs, select subsets of the dataset where only identities of one specific bias type (race-color / religion / gender) are mentioned. We further limit our subset to samples, where exactly one identity is mentioned and where the majority of annotators agreed (2/3 majority) regarding both the toxicity and the identity label to avoid ambiguities.

\subsubsection{BIOS}
\cite{biosbias} proposed the BIOS dataset as part of a case study to investigate gender bias in the context of recruitment systems. The dataset consists of short biographies, which are labeled according to the person's occupation title and gender. 
Due to an automated approach of data retrieval, the dataset quality has been criticized \cite{iwann23schroeder}.
To account for these quality issues, we use the supervised subset of the BIOS dataset provided by \cite{iwann23schroeder}. We also follow their approach and model the task as multi-label classification.

\subsubsection{CrowS-Pairs}
\cite{crowspairs} proposed CrowS-Pairs, a collection of sentences with some (anti-)stereotypical content w.r.t.\ to a protected group paired with a counterfactual version, to evaluate the biases of masked language models. Section \ref{sec:rel_pll} provides an example and explaines the PLL score, with which they measure bias. 
The dataset quality is criticized in \cite{blodgett2021stereotyping}. For instance, they found samples where the counterfactual differed by more than one attribute or samples where race was mixed with nationality or religion.
To address this issue, we only allowed samples with modified tokens (which identify stereotypically associated groups) that are among some predefined terms (refering to \ref{sec:setup_cosine}) and verify that the counterfactual sentence only differs by one bias type. Thereby we excluded both invalid samples (according to \cite{blodgett2021stereotyping}) and those were names are among the modified tokens. Furthermore, we restricted our experiments to the bias types mentioned in Table \ref{tab:protected_groups}, for which a sufficient number of samples was available.

\subsection{Experiment Setup}
\label{sec:exp_setup}

\begin{table}[t]
\caption{Overview of protected groups used in the experiments.}
\begin{center}
\begin{tabular}{ccc}
\toprule
\textbf{Attribute} & \textbf{Groups} & \textbf{\# Defining Sets}  \\
\midrule
race & white, black, asian & 15 \\
religion & christian, muslim, jewish & 14 \\
 & hindu, buddhist &  \\
gender & male, female & 25 \\
age & young, old & 8 \\
\bottomrule
\end{tabular}
\label{tab:protected_groups}
\end{center}
\end{table}

\begin{table}[b]
\caption{Hyperparameters in the classification experiments}
\begin{center}
\begin{tabular}{c|c}
\toprule
cross val folds & 5 (BIOS) / 4 (Jigsaw) \\
optimizer & RMSprop \\
loss & BCEWithLogitsLoss \\
learning rate & 5e-03, 1e-03, 5e-04, 1e-04, 5e-05, 1e-05, 5e-06 \\
epochs & 5 \\
\bottomrule
\end{tabular}
\label{tab:clf_setup}
\end{center}
\end{table}

\subsubsection{Setup for cosine scores}
\label{sec:setup_cosine}
Our experiments cover four protected attributes: race-color, religion, gender and age, which were included in sufficient numbers in at least one of the datasets. Table \ref{tab:protected_groups} shows the protected groups and the number of defining sets. These defining sets were used as attributes and removed from the target samples before computing cosine scores. These attributes were chosen based on the modified tokens in the CrowS-Pairs dataset, so we were able to perfectly remove protected attributes there. In the BIOS dataset a gender-scrubbed version of each biography was available (removing names and pronouns). In addition we removed the additional gender attributes from our defining sets (e.g.\ mother/father). In the Jigsaw dataset we removed terms from the defining sets referring to the labeled group. We assume that our approach doesn't remove all mentions of protected attributes, which would be obvious to a human reader.

Since WEAT is not defined for more than two groups of attributes, we use gWEAT in those cases. 

\subsubsection{Adapting the Direct Bias for multi-dimensional bias spaces}
The Direct Bias was proposed with a one-dimensional bias direction, the 1st Principal Component (PC) \cite{bolukbasi}. However, for protected attributes with more than two groups, it may be more suitable to use a higher dimensional bias subspace as they proposed for their debiasing algorithm, using the first $k$ PCs. Hence, we use:
\begin{align}
    DirectBias(W) := \frac{1}{|T|} \sum_{\mathbf{t} \in T} \left( \sqrt{ \sum_{i = 0}^{|B|} \cos(\mathbf{t},\mathbf{b_i})^2} \right) ^c 
\end{align}
where $B = (b_1, ..., b_k)^T$ is the bias subspace consisting of the first $k$ PCs. For an experiment with $n$ protected groups, we choose $k = n-1$ for a fair comparison with SAME.

\subsubsection{TPR Gap extension}
To apply the TPR Gap (see Section \ref{sec:gap}) for multi-attribute bias, we apply the TPR Gap as the variance of true 
positive rates over all group instead of the difference between two groups. Furthermore, for comparison with the cosine scores, we had to aggregate the TPR Gap over all classes on the BIOS dataset. We did so by computing the mean of absolute class-wise biases.

\subsubsection{MLM Experiment}
We employ $26$ pretrained models from Huggingface with an MLM head. We report the sample-wise cosine scores after removing modified tokens along with  $PLL(S_{more}) - PLL(S_{less})$ of the original sentence $S_{more}$ and its counterfactual $S_{less}$, as well as the aggregated cosine scores and the percentage of samples where $PLL(S_{more}) > PLL(S_{less})$.
Noteworthy, the PLL bias aligns well with WEAT in the sense that both measure specific stereotypes in society. On the other hand, (aggregated) SAME and Direct Bias cannot distinguish whether biases align with the "direction" of a stereotype or its inverse. Hence, we expect WEAT to perform better in comparison with the PLL bias.

\subsubsection{Classification Experiment}
For the classification experiment we use $22$ pretrained models from Huggingface to compute embeddings. We train a separate classification head consisting of two linear layers with ReLU activation in between and a sigmoid activation on top. The hidden size is chosen equal to the embedding size.
To enhance model/ bias variety, we apply the Hard Debiasing Algorithm \cite{bolukbasi} to the pretrained embeddings and train a classification head for the unmodified and debiased embeddings with $k \in {1,3}$ (removing the first k principal components during debiasing \cite{bolukbasi}), resulting in $66$ different classifiers. Additionally, we conduct the experiments via cross-validation, i.e.\ training and evaluating biases on each test fold and reporting the mean biases over all folds.
For the Jigsaw dataset, we handle each protected attribute separately, i.e.\ creating a subset for each protected attribute and training models on these to limit the effects of intersectional bias.
Occasionally, we had issues with some models not converging to a reasonable performance. Hence, we evaluate the classifiers general performance and repeat the training if the model is not suited for bias evaluation (i.e.\ TP rate equal for all classes or recall too low).
Table \ref{tab:clf_setup} shows the hyperparameters and settings used in these experiments. Batch sizes were chosen "as large as possible" on a Nvidia A40 in the range of $1-64$ depending on the models. For GPT models we used a batch size of $1$. The ideal learning rate for each model was chosen by evaluating those given in Table \ref{tab:clf_setup} on a validation set using the F1-score.
The target groups for WEAT were selected by sorting all classes based on their occurrence with the protected groups in the datasets and then assigning all samples with these classes to the respective target groups. In the Jigsaw dataset, it was necessary to assign at least one class to multiple groups, since we had a binary classification, but up to $5$ groups.

\section{Results and Discussion}
\label{sec:results}

\subsection{Predicting downstream bias with pretrained models}
\label{sec:result_predict}

\begin{table*}[t]
\caption{Accuracy of cosine scores of predicting which of two models is more biased. The ground truth is the TPR Gap for classification (BIOS and JIGSAW) and the probability of stereotypical contexts being more likely in MLM (CrowS-Pairs).}
\begin{center}
\begin{tabular}{lccccccccc}
\toprule
\textbf{Cosine} & \textbf{BIOS} & \multicolumn{3}{c}{\textbf{JIGSAW}} & \multicolumn{4}{c}{\textbf{CrowS-Pairs}} & \multirow{2}*{\textbf{avg.\ Rank}} \\ 
\textbf{Score} & \textbf{\textit{Gender}} & \textbf{\textit{Race}}& \textbf{\textit{Religion}}& \textbf{\textit{Gender}} & \textbf{\textit{Race}}& \textbf{\textit{Religion}} & \textbf{\textit{Gender}} & \textbf{\textit{Age}} & \\
\midrule
SAME & $\mathbf{0.66}$ & $\mathbf{0.61}$ & $\mathbf{0.89}$ & $\mathbf{0.58}$ & 0.19 & $\mathbf{0.83}$ & 0.17 & $\mathbf{0.75}$ & 1.4 \\ 
WEAT & 0.51 & - & - & 0.53 & - & - & $\mathbf{1.0}$ & 0.0  & \multirow{2}*{2.3} \\ 
gWEAT & - & 0.59 & 0.64 & - & 0.47 & 0.67 & - & - & \\
Direct Bias & 0.51 & 0.51 & 0.36 & 0.57 & $\mathbf{0.56}$ & $\mathbf{0.83}$ & 0.0 & 0.5 & 2.1 \\ 
\bottomrule
\end{tabular}
\label{tab:bias_pred}
\end{center}
\end{table*}

For experiment \ref{sec:discr_models}, Table \ref{tab:bias_pred} shows the accuracy of ranking two language models in terms of their downstream bias, based on the cosine scores on the pretrained embeddings. We reported either WEAT or gWEAT depending on the number of groups.
In most cases SAME surpasses the other cosine scores. Only in the race and gender experiment with CrowS-Pairs its predictions are completely wrong. Interestingly, in the gender experiment, WEAT perfectly predicts the most/least biased models. A possible explanations is that both WEAT's effect size and the downstream bias measured on CrowS-Pairs measure the exact same stereotypes, while SAME and DirectBias compute mean associations, no matter if they align with the stereotype or oppose it. In the other binary experiment (age) however, WEAT's predictions are reverse to the downstream bias. This could indicate that the MLM heads likelihood is inverse to the association in the embedding space. However, when using all models, we do not observe statistical significant correlations of WEAT and the downstream bias, so this is purely speculative.
Most notably, in both religion experiments SAME (and Direct Bias in the CrowS-Pairs religion experiment) achieve very high scores compared to experiments with the other protected attributes.

\subsection{Correlation of cosine scores with downstream bias}

For experiment \ref{sec:corr_cos}, we investigate the correlation of cosine scores with the downstream bias on three different levels: We compare the aggregated bias scores (one per model), which were also used in Section \ref{sec:result_predict}. As expected, we rarely observe statistical significant correlations. Furthermore, we compare sample-wise biases on the CrowSPairs dataset, which includes counterfactual samples, and we compare class-wise biases on the BIOS dataset, which is multi-class classification.

\subsubsection{Class Biases}
\begin{figure}[t]
\centerline{
\includegraphics[width=0.25\textwidth]{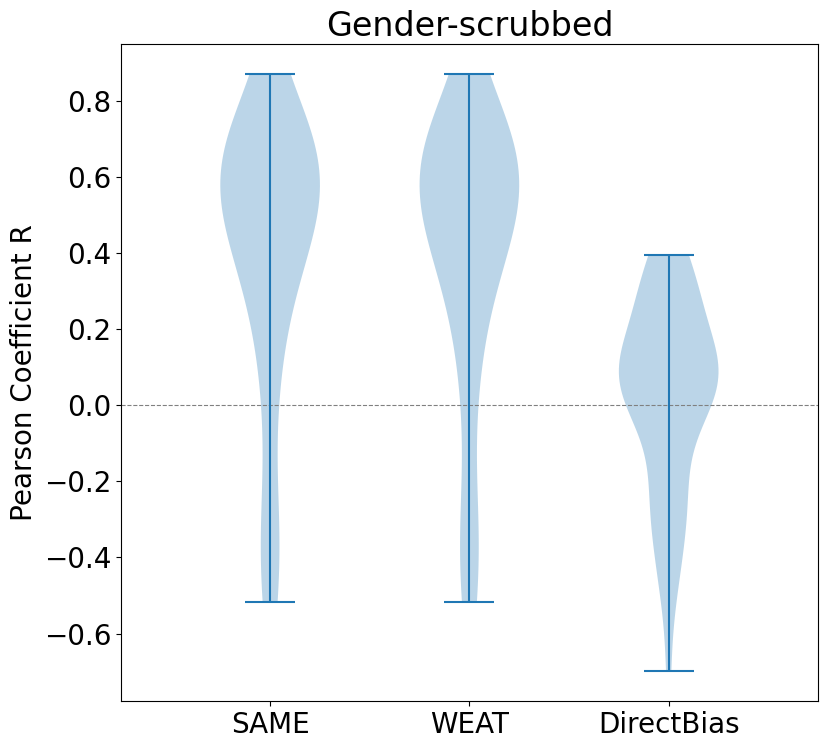} 
\includegraphics[width=0.25\textwidth]{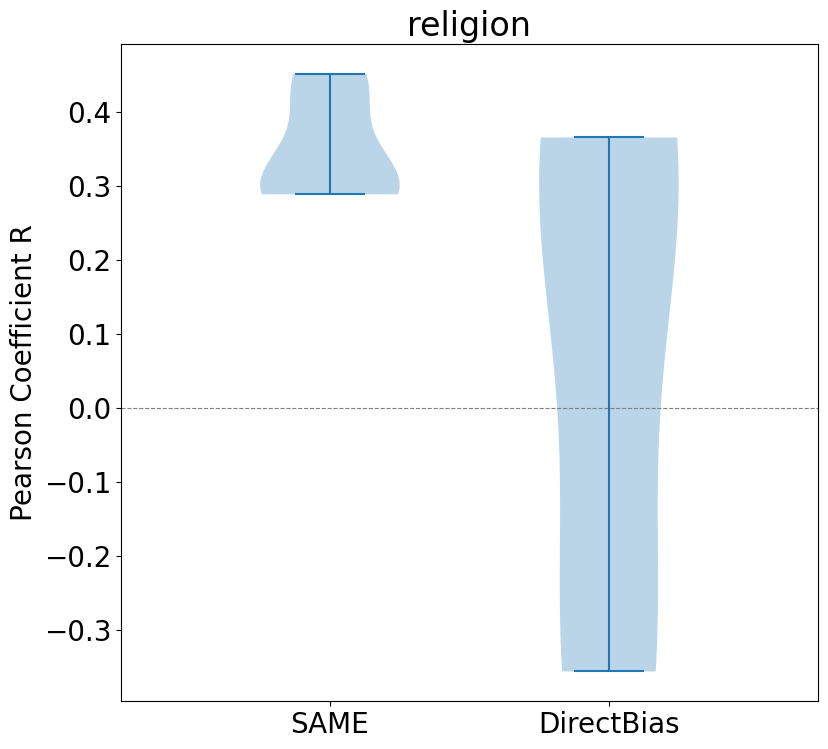} 
}
\caption{Distribution of Pearson Coefficient R for class-wise biases and sample biases. Left: Correlation of class-wise cosine scores and TPR Gap on the BIOS dataset with gender-scrubbed data. Right: Correlation of sample-wise cosine scores with the PLL difference on the CrowS-Pairs dataset for religion bias, including those models with $p > 0.05$. 
}
\label{fig:bios_class_bias}
\end{figure}

Figure \ref{fig:bios_class_bias} (left side) shows the correlation of class-wise biases on the BIOS dataset. The class-wise biases refer to the class-wise TPR Gap as defined in \ref{sec:gap} and the mean sample biases per class of the cosine scores. Although the aggregated cosine scores did not correlate, we see a connection between the mean gender association (WEAT and SAME) of a classes samples with the downstream TPR Gap score, i.e.\ classes, whose gender-scrubbed samples were more similar to female attributes, also tended to achieve a higher true positive rate for the female samples. For the Direct Bias we achieve pearson coefficients distributed around 0, but for WEAT and SAME we report higher correlations for a majority of models. We observed similar results when training on raw BIOS (including gender indicators), but slightly more models with low Pearson Coefficients. Presumably, a classifier trained on the raw texts would simply pay attention to gendered pronouns as opposed to semantic bias when making predictions correlated to gender. However, when trained on gender-scrubbed data, the classifiers seem to rely even more on the semantic bias.

\subsubsection{Sample Biases}
We evaluate sample biases on the CrowS-Pairs dataset, where we have actual counterfactual samples. For the majority of models, we do not observe a statistical significant correlation of the PLL difference and the cosine scores. Figure \ref{fig:bios_class_bias} (right side) shows the results for the religion experiment, where we observe some models with statistical significant correlations ($p < 0.05$). Considering these, we report a wide range of Pearson coefficients for the Direct Bias while SAME consistently achieves $R$ around $0.3$ to $0.4$. This implies a connection between the semantic bias measured by SAME and the downstream bias. However, it also shows that different models may encode biases differently.


\subsubsection{Aggregated Bias}

\begin{figure}[t]
\centerline{
\includegraphics[width=0.25\textwidth]{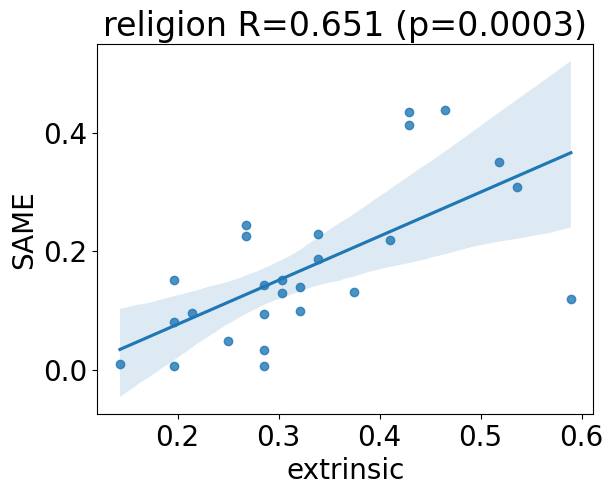}
\includegraphics[width=0.25\textwidth]{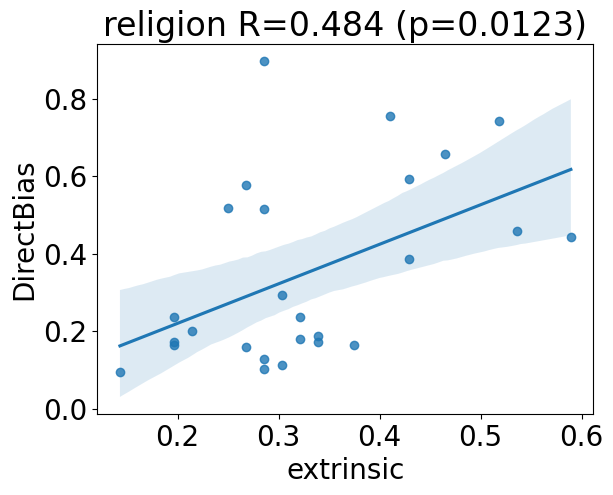}
}
\caption{Correlation of aggregated cosine scores with the extrinsic bias in the CrowS-Pairs dataset (\% of samples where stereotypical version was more likely according to the language model).} 
\label{fig:mlm_religion_aggregated}
\end{figure}

We test the correlation of aggregated bias scores with the downstream bias on all three datasets, but mostly find no statistical significant correlations. One exception is the experiment on religion bias in the CrowS-Pairs dataset. Figure \ref{fig:mlm_religion_aggregated} shows the correlation plots of SAME and the Direct Bias with the PLL bias. Both achieve rather high Pearson Coefficients with $p < 0.05$, although SAME achieves a better correlation. Considering the results in Table \ref{tab:bias_pred}, where SAME and the Direct Bias performed equally well, this shows that SAME is overall more accurate.

\subsection{Robustness of bias scores}
\begin{figure}[t]
\centerline{
\includegraphics[height=0.17\textwidth]{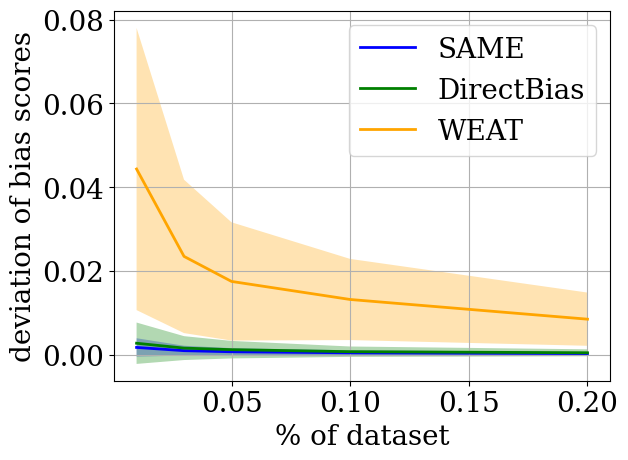}
\includegraphics[height=0.17\textwidth]{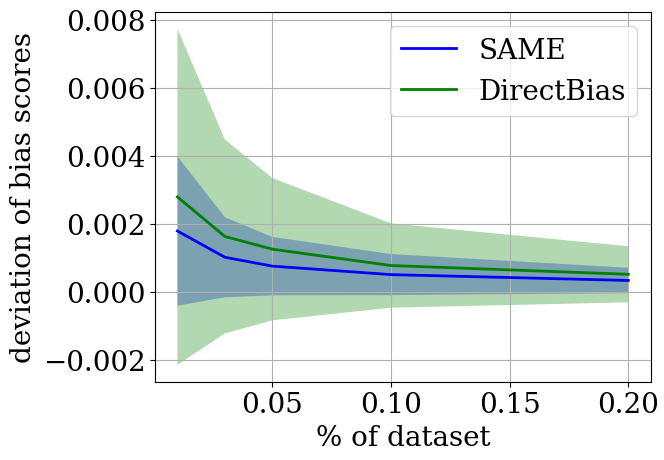}
}
\caption{Deviation of bias scores when computed on smaller subsets of the BIOS dataset (of a total of 10602 samples). Smaller values indicate higher robustness. Right plot excludes WEAT for better resolution}
\label{fig:robustness_target}
\end{figure}
Figure \ref{fig:robustness_target} shows the deviation of bias scores when computed on smaller subsets of the BIOS dataset. We start with 20\% (2120 samples) of the data, which matches our test set size in the other experiments and decrease to 1\% (106 samples). As we can see in the figure, SAME and the Direct Bias are more stable for smaller subsets. This is expected, since WEAT's partitioning of test samples into stereotypical groups also depends on the test data in our experiment. Especially for the smallest subsets, we can expect class-gender distributions to vary significantly. A well informed hypothesis/ manual design of stereotypical groups should be able to mitigate this effect to some extent. When zooming in on SAME and the Direct Bias, we observe that SAME is the most robust both in terms of overall deviations and the standard deviation, although differences are rather low.

We further compared the bias measures computed on job titles (single words) and on 50\% of the bios (job titles in context). We reported the following absolute differences of bias scores, where WEAT's scores were normalized to the interval [0,1] for better comparison. For SAME we reported bias score differences of 0.0454 (+/- 0.0368), for WEAT 0.1126 (+/- 0.0779), and for the DirectBias 0.1135 (+/- 0.1141).
Both tests shows that SAME is more robust to the selection of targets, including the simplified case, where biases are computed based on single words without context. This experiment also highlights the importance of choosing a sufficient amount of data and contextualized samples over single words. When approaching the amount of test data used in the previous experiments (20\% of bios) we observe only minor  deviations of cosine scores. This implies that the reduced BIOS dataset (see Section \ref{sec:exp_datasets}) and the amount of test data chosen in the previous experiments were a proper choice.

\section{Conclusion}
\label{sec:conclusion}

In this paper, we proposed SAME, a novel bias score for semantic bias in pretrained language models. Using theoretical criteria from the literature, we could show that SAME is better suited to quantify social bias since it overcomes limitations of similar bias scores. Our experiments confirm that SAME is best suited to rank more/less biased models. We observed few cases, where WEAT performs very well. This makes sense, since one of our baseline scores tested exactly the same hypothesis as WEAT, so they should produce similar results. However, we have no further results (actual correlation) to substantiate this.
As expected, statistical significant correlations of bias scores with downstream bias were rather scarce, presumably because other bias causes (e.g. in the datasets or downstream heads) had a stronger influence. However, we found some cases where cosine scores, most notably SAME, did correlate with the downstream bias. While these correlations may not be extremely high, even moderate correlations of semantic and downstream bias show that we cannot neglect this cause for bias and, according to our results, SAME is the most promising of the observed measures for this. Considering the class bias correlations, it becomes clear that only looking at aggregated bias scores may not be the ideal solutions. In a classification task, the biases regarding the relation of classes may be more revealing than an aggregated score. Finally, SAME also performed well in terms of robustness.

In conclusion, we showed both theoretically and empirically that SAME is better suited to quantify bias. There may be an exception for use cases where a specific hypothesis should be queried, but future work should back up this claim by proofs or substantial empirical evaluation. Furthermore, it would be interesting to compare SAME to other non-cosine intrinsic biases which could also be applied to evaluate social bias in pretrained language models.

\section*{Acknowledgment}
Funded by the Ministry of Culture and Science of North-Rhine-Westphalia in the frame of project SAIL, NW21-059A.

\end{document}